# Controllable 3D Generative Adversarial Face Model via Disentangling Shape and Appearance

Fariborz Taherkhani[1,2], Aashish Rai[1*], Quankai Gao[1*], Shaunak Srivastava[1*], Xuanbai Chen[1], Fernando de la Torre[1], Steven Song[2], Aayush Prakash[2], and Daeil Kim[2]

[1]Carnegie Mellon University
[2]Facebook/Meta
[*]*(equal contribution)*

## Abstract

*3D face modeling has been an active area of research in computer vision and computer graphics, fueling applications ranging from facial expression transfer in virtual avatars to synthetic data generation. Existing 3D deep learning generative models (e.g., VAE, GANs) allow generating compact face representations (both shape and texture) that can model non-linearities in the shape and appearance space (e.g., scatter effects, specularities,..). However, they lack the capability to control the generation of subtle expressions. This paper proposes a new 3D face generative model that can decouple identity and expression and provides granular control over expressions. In particular, we propose using a pair of supervised auto-encoder and generative adversarial networks to produce high-quality 3D faces, both in terms of appearance and shape. Experimental results in the generation of 3D faces learned with holistic expression labels, or Action Unit (AU) labels, show how we can decouple identity and expression; gaining fine-control over expressions while preserving identity.* [1]

## 1. Introduction

Photo-realistic 3D face generation has sparked a lot of interest in the domain of computer graphics, and computer vision, fueled by applications such as creating virtual avatars [36], face recognition [4, 19], 3D face animation [26, 54], expression transfer [29, 50, 41], and using generative models to create synthetic training data [46] to improve the performance of down-stream tasks in computer vision such as 3D face reconstruction [20]. Computer graphics or machine learning generative models (or a combination of two) can create synthetic 3D faces and both have their own set of advantages and disadvantages [47, 57, 51]. In general, physics and anatomical models with firm control over expression (e.g., blend shape), camera position, skin texture, or lighting assisted production of 3D faces is done using computer graphics. However, for these computer graphics models to work well, access to high-quality assets is required, which often necessitates a significant amount of artistic labor, both costly and time-consuming. However, approaches based on generative models (e.g., GANs [22], VAE [30]) learned from visual data can generate instances of 3D faces automatically and provide photo-realistic models with natural image statistics useful for many applications [44, 59, 15, 36]. They do, however, need a sufficient amount of well-balanced data to learn from, and model training to achieve fine-grained control over features of interest like subtle expressions, skin tone, or lighting which is sometimes more challenging.

This paper addresses the issues of granular control of 3D faces using generative models. Specifically, this work proposes a new 3D shape and appearance generative model that can synthesize high-quality 3D faces with granular control over expressions. Fig. 1 illustrates the capabilities of the model. The model can decouple identity and expression factors and generate fine-controllable expressions of a given identity in both shape and UV texture domains. The 3D generative model must overcome several challenges to reach these capabilities. First, the model has to decouple identity and expression. Maintaining the identity while varying expression is critical for applications such as facial expression transfer [39, 41]. We achieve this by providing high-level labels in the training set as holistic expression labels or action units. Recall that manually labeling the intensity of expression consistently is highly challenging and time-consuming. However, our method can synthesize expressions over a range of intensities (see Fig. 1.(c,d)). Second, the data's high-dimensional input (e.g., tens of thousands vertices with their 3D coordinates for each mesh) combined with relatively small-sized training data can lead to over-fitting and lack of generalization. To address this issue, we propose a supervised auto-encoder (SAE) to find compact and discriminative representations for expressions and identities in a statistically meaningful way. This approach constructs regions in the identity and expression latent space where similar data points come together and cluster. This results in a simplified sampling procedure for generating 3D faces. Fig. 1.(a) shows how our method decouples identity and expression. Third, modeling the complex distribution of the data represented by the SAE and estimating how they are generated in a probabilistic framework is not a trivial task. Our framework adopts a conditional GAN (cGAN) [38] that learns to sample from the disentangled subspaces of the SAE. Here, we chose cGAN as we aim to control the type and intensity of expressions when generating new identities. Similarly, we use another cGAN trained on high-resolution texture maps to synthesize facial appearance as our statistical representation of the facial texture. See Fig. 1.(b) and (c) for the generated UV texture maps and rendered photo-realistic 3D faces, respectively.

---

[1]https://aashishrai3799.github.io/3DFaceCAM/

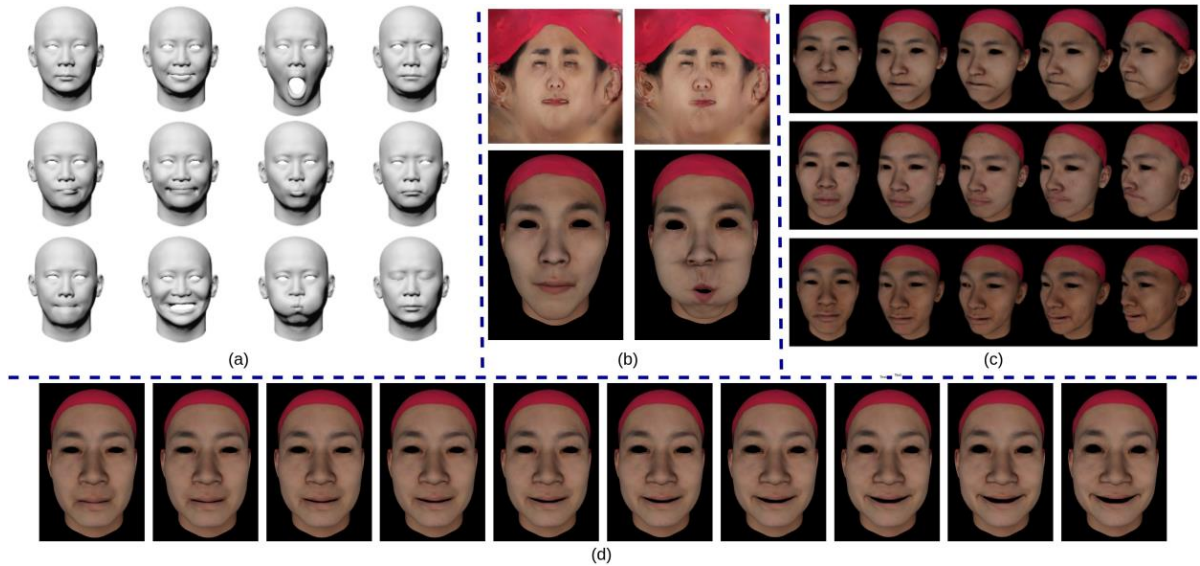

Figure 1. Our generator's uncurated set of shapes, textures and rendered faces with the FaceScape dataset[56]. (a) Shapes of different expressions belonging to the same identity. (b) Expression-specific generated textures and corresponding rendered faces. (c) Each row shows multi-view extrapolation of the expression intensity while preserving the identity. (d) Facial expression (Smile) synthesis with different monotonic intensity.

## 2. Previous Work

Active appearance [13, 21, 17], kernel models [24], 3D Morphable Models (3DMM) [6], and their deep learning extensions [15] are common models for data-driven 3D face synthesis. In the early 90s', 3D face modeling via 3DMM was a common practice for its power in compact representation and providing a strong prior on the shape variations with respect to their natural factors (e.g., identity and expression). The original 3DMM [6] disentangles geometry, expression [10], and colored texture using PCA models. These models and their variants [9, 34, 53, 7, 8, 10, 34] are some of the most well-known approaches for modeling facial textures and shapes independently. However, PCA and its variants are linear and cannot effectively capture high-frequency signals. Thus, it is hard for 3DMM to model the subtle differences in facial shapes and textures with linear models.

In contrast to linear models, kernel and deep-learning methods can nonlinearly model the variability of shape and texture. Tan et al. [49] proposed a method based on VAE [49, 5, 30] to efficiently compress and represent several classes of 3D shapes. The idea here is to model the deformation of meshes in a local coordinate frame [35] and reconstruct the positions of the mesh using a different linear model. Ranjan et al. [43] used non-linear models based on convolutional mesh autoencoders and graph convolutions to improve the expressiveness of face geometry. Even though these models can achieve better reconstruction than linear models, disentangling facial identity and expression is overlooked. Several works [2, 5, 3, 19, 27] have focused on disentangling facial identity and expression in the architecture of the network. The state-of-the-art works in this line of research include [10, 34]. While most of these approaches learn to implicitly disentangle identity and expression, some other methods explicitly include the disentanglement in the design of style, and architecture [37, 41, 11, 33]. While

[11, 33] are similar to each other $w.r.t$ the architecture design, [11] additionally decouples the identity and expression in the latent spaces, and use a so-called joint decoder to model identity-specific expression deformations.

Recently, GANs have been used for 3D face representation, generation and expression style transfer [39, 41]. Even though [39] obtains decent results for reconstruction, it is not able to transfer the expressions properly. This is because the identity and expression factors are not explicitly decoupled due to a shared latent space. However, [41] addresses this issue by using image-to-image translation networks in the 3D domain. [14] first fits a 3DMM to images and then applies GAN to complete the missing parts obtained from their UV maps. On the other hand, [25, 55] take a 3D face as input and then learn to improve its geometry using photometric information via a GAN. [48, 46] learn identity variations by training a GAN on the UV-maps. These methods ignore to model the non-linear variations and the correlation existing between identity and expression. However, [2] considers the non-linear variations and this correlation and then decouples them using the GAN.

Besides modeling 3D facial shapes, GANs have also been used to generate 3D facial textures [48]. [20] replaces 3DMM with a GAN to reconstruct the texture while keeping the statistical models to reconstruct the shape. [31] uses an image-to-image translation technique using GANs to generate per-pixel diffuse and specular components from a texture map. However, [39] models 3D shapes using GANs in a parametric UV map. While [39, 43] disregard the correlation between shape and texture, [19] considers the correlation and normal map to generate high-fidelity 3D facial images. On the other hand, [45] and [18] use style transfer GANs for generating photo-realistic images of 3DMM.

## 3. Method

This section describes the proposed generative model for 3D shape and appearance. The overview of our 3D generative model is shown in Fig. 2.

### 3.1. 3D Shape modeling

The shape component is generated in two steps, the Supervised Auto-Encoder(SAE) and the GAN.

**In the first step,** as shown in Fig.2, we train an SAE, which projects shapes into two low dimensional embedding subspaces, one of which is dedicated to capture the identity factor while the other is used to capture the expression factor. The SAE contains two encoders that share no parameters, allowing us to decouple and disentangle the representation of the identity and expression factors. However, the SAE shares the same decoder that supervises the two encoders, to preserve the correlation between the identity and expression factors while reconstructing the mesh.

The SAE uses the classification loss on identity/expression classes as a supervision signal. Precisely, this supervision separates classes of expressions and identities and brings similar expression and identity factors together in the embedding spaces. This gives us a prior on the identity and expression embedding spaces and simplifies the sampling procedure.

Let $x$ be the original input mesh, with $\mu_{exp} = \mathcal{E}_e(\theta_e, x)$ and $\mu_{id} = \mathcal{E}_i(\theta_i, x)$ being the expression and identity encoders with parameters of $\theta_e$, and $\theta_i$ which take $x$ as input and project it into the identity and expression subspaces, respectively. Let $\mathcal{D}(\theta_d, \mu)$ be the decoder with parameters of $\theta_d$ which takes $\mu$ as input where $\mu = (\mu_{id}, \mu_{exp})$ is the concatenation of the two encoders' outputs. The objective function for training the SAE including parameters $\theta = \{\theta_e, \theta_i, \theta_d, w_e, w_i\}$ is:

$$\begin{aligned}\mathcal{L}_{SAE}(\theta, x) =& ||x - \mathcal{D}(\theta_d, \mu)||_1 + \mathcal{L}_c(w_i \mathcal{E}_i(\theta_i, x), y_i) \\ &+ \mathcal{L}_c(w_e \mathcal{E}_e(\theta_e, x), y_e),\end{aligned} \quad (1)$$

where $\mathcal{L}_c(.)$ denotes the softmax cross-entropy, $y_i$ and $y_e$ are the one-hot identity and expression labels, respectively. $w_e$, and $w_i$ are the parameters of the identity and expression classification layers, respectively.

To show the superiority of our SAE, we compare its identity and expression embedding spaces against an unsupervised AE, with the same architecture as the SAE. The identity and expression embedding spaces are computed with t-SNE [52] as shown in Fig. 3. As expected, in Fig. 3(a) and (c), both expression and identity embeddings can be seen clustered when using SAE. There are 20 clusters in the expression embedding space, and each cluster includes 847 different identities per expression as shown in Fig. 3(a). Fig. 3(c) shows the identity embedding space, which has 847 clusters and each cluster includes 20 different expressions per identity. However, both expression and identity embedding spaces are mixed and noisy when using unsupervised AE, as shown in Fig. 3(b) and Fig. 3(d), respectively. Note that there are 20 expressions for each of the 847 identities in the dataset used in this experiment. To see experiments about t-SNE on testing samples please refer to Appendix 1.

**In the second step,** as shown in Fig. 2, we leverage a conditional GAN (cGAN) framework to sample from the distribution of the identity and expression factors represented in the embedded spaces obtained in the first step. The cGAN learns a mapping from an input $y$ and $z$ to the output $\mu : G(y,z) : \{y,z\} \rightarrow \mu$. In this paper, we use a cGAN to learn a mapping function from $z$, and expression/identity class label codes to real data from the disentangled subspaces learned from the SAE. We use cGAN as we want to control the type of expressions when generating new identities. Specifically, the cGAN takes as input a vector $(z_{id}, z_{exp}, z_{noise})$ which is the concatenation of the identity code $z_{id} \sim \mathcal{N}(0, 1)$ of dimension $n_{id}$, expression code $z_{exp} \sim p_{exp}$, where $p_{exp}$ denotes the distribution of the expression classes of dimension $n_{exp}$, and $z_{noise} \sim \mathcal{N}(0, 1)$. Similar to SAE, we use two separate generators for identity and expression factors ($\mu'_{id}$, $\mu'_{exp}$) as we aim to decouple them during the generation of the fake samples. However, we use a shared discriminator which takes concatenation of the two generators' outputs, ($\mu' = (\mu'_{id}, \mu'_{exp})$) and tends to preserve their correlation when generating the fake identity and expression embedding.

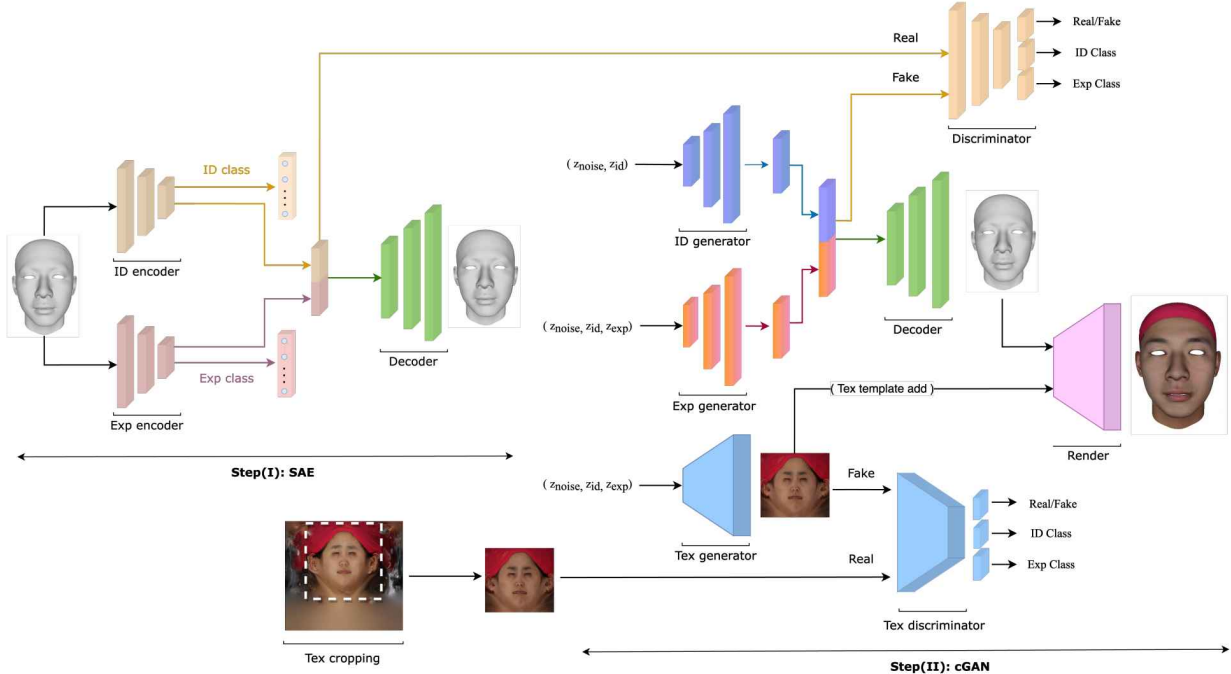

Figure 2. Overview of our 3D generative model. The first step includes training an SAE, which projects shapes into two low dimensional embedding subspaces, one of which is dedicated to the identity factor while the other to the expression factor. In the second step, we utilize a cGAN network to sample shape and texture from the distribution of the identity and expression factors. A renderer is then used to generate photorealistic faces.

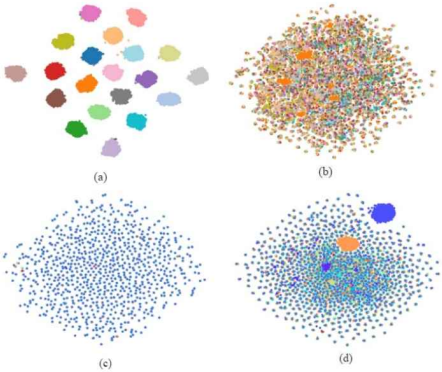

Figure 3. Visualization of embedding spaces on FaceScape dataset [56] by t-SNE: (a) expression embedding using SAE, (b) expression embedding using unsupervised AE, (c) identity embedding using SAE, and (d) identity embedding using unsupervised AE. There are 20 colors in each figure indicating 20 expressions.

We also add identity and expression classification losses in the adversarial loss so that the discriminator returns the probability of a mesh representation belonging to a pre-defined class label, as it benefits the GAN's performance [40]. This loss encourages both the generator and the discriminator to distinguish whether two expressions or identities with the same embedding code (i.e., $z_{id}$, $z_{exp}$) are perceptually similar. Moreover, these losses cause our model to decouple the identity and expression factors further because the classification of one factor (e.g., identity) is independent of the choice of the labels of the other factor (e.g., expression). In our case, the two generators take both the $z_{noise}$ and corresponding expression/identity class label codes. Thus, the identity generator takes $z_{noise}$ and $z_{id}$, and returns the fake identity code: $\mu'_{id} = G_{id}(z_{noise}, z_{id})$ and the expression generator takes $z_{noise}$, $z_{id}$ and $z_{exp}$, and returns the fake expression code: $\mu'_{exp} = G_{exp}(z_{noise}, z_{id}, z_{exp})$. Adding $z_{id}$ to the expression generator helps control identity-specific fine details. On the other hand, the discriminator gives probability distribution ($s$) over the source data which is either real $\mu = (\mu_{id}, \mu_{exp})$, or fake $\mu' = (\mu'_{id}, \mu'_{exp})$ and the probability distribution over the expression ($c_{exp}$) and identity ($c_{id}$) class labels which are, $p(s|\mu)$, $p(c_{id}|\mu)$, and $p(c_{exp}|\mu)$, respectively. Therefore, our complete loss function for training each cGAN contains two terms: log-likelihood of the

correct source, $\mathcal{L}_s$, and log-likelihood of the correct class label for expression, $\mathcal{L}_{exp}$ and identity, $\mathcal{L}_{id}$.

$$\mathcal{L}_s = \mathbb{E}_{\mu \sim p_{data}(\mu)}[\log p(s=1|\mu)] + \\ \mathbb{E}_{z \sim p_z(z)}[\log p(s=0|\mu')], \quad (2)$$

where $s = 1$, and $s = 0$ denote the label of real and fake data, respectively. Here, $z$ denotes concatenation of $z_{noise}$, and corresponding expression/identity class label codes $z_{exp}/z_{id}$. $p_{data}(\mu)$ denotes real distribution of data represented by our SAE : $\mu = (\mu_{id}, \mu_{exp})$,

$$\mathcal{L}_{id} = \mathbb{E}_{\mu_{id} \sim p_{data}(\mu_{id})}[\log p(c=c_{id}|\mu_{id})] \\ + \mathbb{E}_{z \sim p_z(z)}[\log p(c=c_{id}|\mu'_{id})], \quad (3)$$

$$\mathcal{L}_{exp} = \mathbb{E}_{\mu_{exp} \sim p_{data}(\mu_{exp})}[\log p(c=c_{exp}|\mu_{exp})] \\ + \mathbb{E}_{z \sim p_z(z)}[\log p(c=c_{exp}|\mu'_{exp})], \quad (4)$$

where the discriminator is trained to maximize $\mathcal{L}_s + \mathcal{L}_{id} + \mathcal{L}_{exp}$, while identity/ expression generators are trained to minimize $\mathcal{L}_s - \mathcal{L}_{id}$ and $\mathcal{L}_s - \mathcal{L}_{exp}$, respectively.

### 3.2. Texture Generation

For texture generation, we used a progressive GAN [28] and conditioned it on the identity and expression code. The shape and texture generators are trained with the same input codes ($z_{id}$, $z_{exp}$, $z_{noise}$). This enables our model to correlate shapes and the corresponding textures and generate them for rendering. See Fig. 2 for details of the architecture.

The input to the progressive generator is a vector with three components, namely ($z_{id}$, $z_{exp}$, $z_{noise}$). Each $z_{id}$ is randomly sampled from a Gaussian distribution and is fixed corresponding to a specific identity class throughout the training. $z_{exp}$ is a one-hot vector that specifies the training sample's expression. $z_{noise}$ is a vector randomly sampled from the Gaussian distribution but varies during training. $z_{id}$ and $z_{exp}$ are kept identical for both the shape and texture generators. The last layer of the discriminator is split into three branches to get 1) real or fake, 2) expression class, and 3) identity class. We used the WGAN-GP [23] loss to train the generator and discriminator in a progressive setting. Similar to cGAN for shape generation, cross-entropy identity/expression classification losses are added to our adversarial loss so that the discriminator returns the probability of the textures belonging to a pre-defined class label for improving the performance. See Fig. 16 in Appendix 1 for synthesized texture maps. There is also a texture pre-processing step including cropping the frontal portion of the face to avoid the redundant noise in the remaining parts of the texture maps, which can better stabilize the training of the texture generator. Thus, the texture generated from our model is first added to the template texture before rendering.

As shown in Fig. 16, $z_{id}$ and $z_{exp}$ allow us to control the identity and expression of the textures. By fixing the $z_{exp}$, we can fix the expression and change the identity by varying the $z_{id}$ and vice-versa. Fig. 17 in Appendix 1 shows an example of meshes and rendered images of a synthesized identity with different expressions generated by our model. Fig. 4 shows synthesized rendered faces from our method using randomly sampled latent codes. It is interesting to note the kind of diversity our method can generate in terms of skin color, age, gender, facial features and face shape.

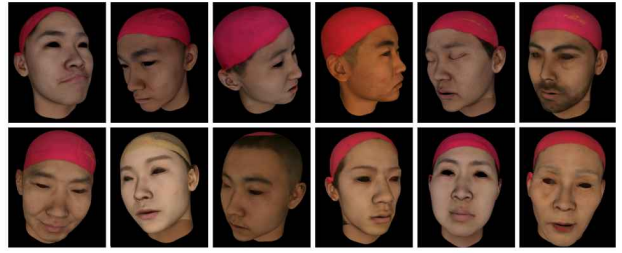

Figure 4. Randomly synthesized faces from our method showcasing diversity in age, gender, skin color, facial features and face shape.

## 4. Experiments

Here, we describe the experimental validation. The datasets, pre-processing, and implementation details are first described. The first experiment shows how we can generate new identities and expressions. Specifically, we show how our method can generate fine-controllable expressions, generate mixed expressions, transfer expressions to new identities, and perform style editing. The second experiment evaluates the quality of the 3D shape generative model using quantitative metrics (diversity, specificity) described in [1]. Finally, we show how our method can be used to synthesize controlled subtle 3D facial expressions.

### 4.1. Datasets

**FaceScape dataset**. We conduct our experiments on FaceScape dataset [56], which is a large-scale 3D face dataset and includes 16940 (with 847 identities and 20 expressions) topologically uniformed 3D face models with displacement maps and 4K high quality texture maps. Each mesh has 26317 vertices with corresponding 3D coordinates.

**BP4D-Spontaneous dataset**. To demonstrate that our model works well with Facial Action Coding System (FACS), specifically with Action Units (AUs), and can handle highly diverse data, we used the BP4D-Spontaneous dataset [58] to conduct further experiments. This dataset is a collection of 41 diverse identities with many spontaneous video frames for 8 tasks. It contains unregistered meshes and textures along with 2D images for each frame and has 34 labeled AUs. We uniformly sampled around 50 frames from each task to get 17790 frames. We then used PRNet [16] to generate frontal face registered meshes and corresponding textures of each frame. Each generated mesh has 43867 vertices with corresponding 3D Cartesian coordinates. The textures are 256x256 resolution.

### 4.2. Implementation details and architectures

**Shape generation:** The input of our identity generator $G_{id}$ is ($z_{id}$, $z_{noise}$) and the input of our expression generator $G_{exp}$ is ($z_{id}$, $z_{exp}$, $z_{noise}$) such that the expression is conditioned on identity (refer Appendix 1 for details). The output of our identity generator is ID embedding vector and the output of our expression generator is Exp embedding vector. Both generators are fully-connected networks. The discriminator $D$ consists of a common branch followed by 3 branches in the last layer. The first branch is to determine whether input sample is real or fake, the second branch is to predict the identity class, and the third branch is to predict which expression class an input sample belongs to.

As for the auto-encoder, the architectures of our identity and expression encoder are both fully-connected networks. The output of identity and expression encoder is 2-fold: one is for identity or expression embedding, the other is to predict the identity or expression class. The architecture of the encoders/decoder, and generators/discriminator are provided in detail in Table 3, 4, 5, 6 in the Appendix, respectively. We render the correlated texture and shape with MeshLab [12], which is an open-source rendering tool.

**Texture generation:** The input of texture generator is an embedding vector ($z_{id}$, $z_{exp}$, $z_{noise}$). A large chunk of texture maps within FaceScape dataset have blurred eyes to preserve the identity. This blurriness around the eyes results in artifacts. To avoid such artifacts, we used only those textures that do not have blurred eyes, which results in 359 subjects totaling around 7k texture maps for training. We progressively [28] trained the model starting from a resolution of $16 \times 16$ and progressing up to $512 \times 512$ (doubling the resolution in each step) to generate textures that can result in photo-realistic rendered images. Our generator and discriminator architectures are similar to Progressive GAN [28]. The only difference is we went up to a maximum resolution of $512 \times 512$ starting with $z_{noise}$ of 256.

### 4.3. 3D shape synthesis

Here, we describe experiments that sample from the identity, expression spaces, and illustrate the semantic properties of our embedding by interpolating the expression space.

**Identity Synthesis:** One of the benefits of the proposed ID generator $G_{id}$ is that it can produce different identities by changing the input code of identity while keeping the input code of Exp generator $G_{exp}$ fixed. Fig. 5 shows examples of different identities generated by our model. Along the identity axis, different identity codes $z_{id}$ are randomly sampled from a Gaussian distribution. The choice of the conditional generator also allows smooth interpolation between two identities by linearly interpolating their identity codes, as shown in Fig. 8.

Table 1. Quantitative metrics of our method w.r.t normalized Diversity(DIV, DIV-ID and DIV-EXP) and absolute Specificity(SP) on Facecape dataset. Higher is better, except for specificity. Level 1,5,10 shows the level of extrapolation for the expressions. *Ohe-hot* and *Gaussian* specifies $z_{id}$ type. SP values are in *mm*.

|  | DIV ↑ | DIV-ID ↑ | DIV-EXP ↑ | SP ↓ |
|---|---|---|---|---|
| Training data | 1 | 1 | 1 | - |
| 3DMM [2] | 0.72 | 0.59 | 0.57 | 2.30 |
| MAE [3] | 0.79 | 0.28 | 0.75 | 2.00 |
| CoMA [41] | 0.69 | 0.52 | 0.58 | 2.47 |
| Victoria et al. [1] | 0.96 | 0.58 | 0.84 | 2.01 |
| *Ours(One − hot)* | | | | |
| Level 1 | 0.77 | **0.81** | 0.37 | **0.84** |
| Level 5 | 0.76 | 0.78 | 1.13 | 0.86 |
| Level 10 | 0.77 | 0.76 | 2.03 | 0.94 |
| *Ours(Gaussian)* | | | | |
| Level 1 | 0.75 | 0.75 | 0.8 | 0.84 |
| Level 5 | 0.86 | 0.74 | 3.83 | 0.86 |
| Level 10 | **1.26** | 0.73 | **7.86** | 0.94 |

**Expression Synthesis:** As shown in Fig. 5, the Exp generator $G_{exp}$ allows us to synthesize shapes with various expressions by varying the expression code $z_{exp}$ while keeping the input code of ID generator $G_{id}$ fixed. Note that in these results, we also show that the expressions can readily be transferred from one identity to another by changing the $z_{id}$ code in the $G_{id}$ while keeping all other values fixed. Similar to the identity space, the model also allows smooth interpolation between two expressions without changing the identity by linearly interpolating their expression codes. Our model can also semantically control the intensity of each expression by extrapolation, analogous to learning its own implicit blendshapes [32], and produces plausible variations in some ranges we defined, which can be seen in Fig. 6.

Our model also makes the interpolation between different expressions possible. Fig. 7 shows the result of superimposing two expressions to produce a new natural looking mixed expression. This feature can also be used for style editing where our model can transfer fine-details associated in a shape's identity by transferring the identity code $z_{id}$ from one identity to another in the expression generator when these two shapes have the same expression code $z_{exp}$. See Appendix 1 for more details and examples.

**Quantitative evaluation of 3D shape synthesis:** The performance of 2D GANs is often measured using the FID score, which is not that meaningful for 3D models. Similar to [1], the quantitative metrics of our model include diversity and specificity of the generated 3D shape samples.

*Diversity:* One of the important indicators about a generative model is to what extent can we generate samples showing enough diversity, which is measured by calculating the mean vertex distance over $n$ pairs of generated samples. The higher value of diversity we get, the more diverse samples our model can generate. Here, we use $n$=10000.

*Specificity:* Diversity alone is not enough because even contorted shapes (irregular human faces) can contribute to diversity significantly. Thus, specificity is provided to measure the distribution shift between the generated data and the original training data. Ideally, the distribution of the generated data is expected to be as close as possible to the distribution of the original training data. To this end, $n$ samples are randomly generated and we calculate the mean value over the mean vertex distances between each generated sample and each member of training set. Here, we use $n$=1000. Comparing results using these metrics are reported in Table. 1.

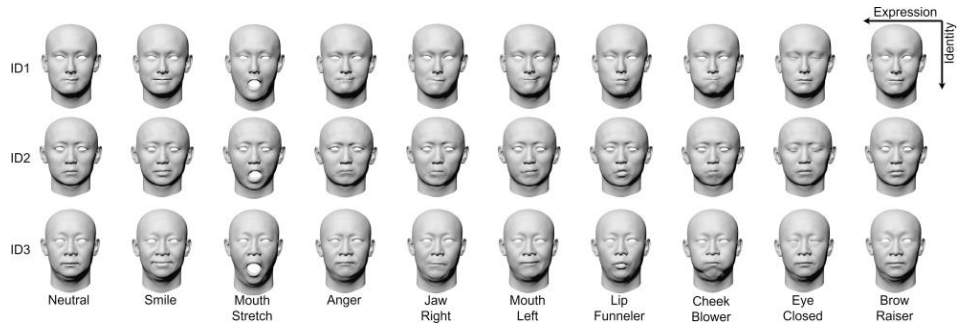

Figure 5. **Novel Identity Synthesis:** Synthesizing a set of novel identities along with the desired expressions by sampling from latent space.

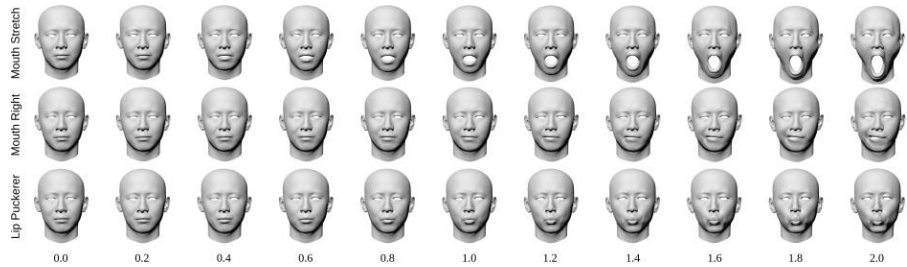

Figure 6. **Varying intensity of Expressions by Extrapolation:** Faces show smooth increase in expressiveness as we vary the intensity along the expression dimension.

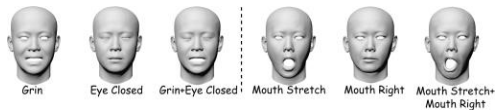

Figure 7. Superimposing expressions with mixed expressions.

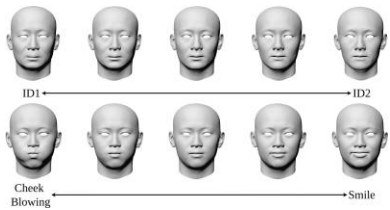

Figure 8. Smooth linear interpolation across identity (top) and expression (bottom).

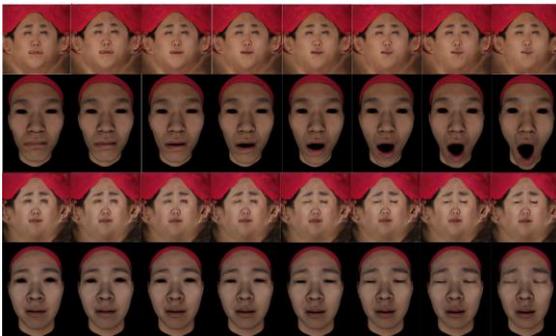

Figure 9. Texture intensity variation along with rendered images.

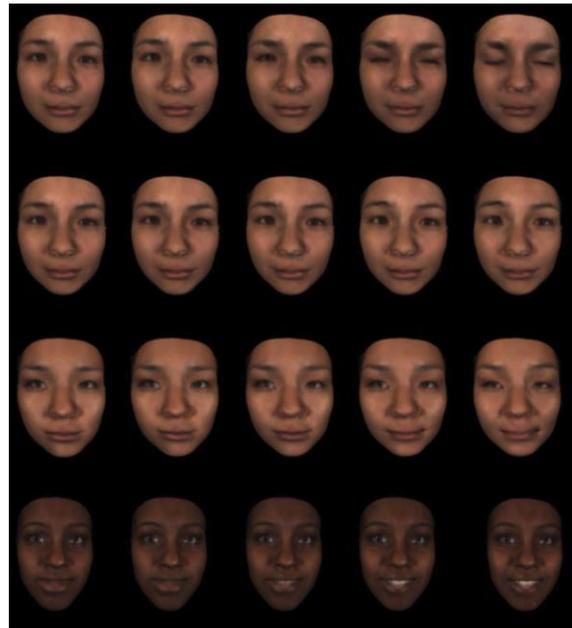

Figure 10. We show that our model works well with a highly-diverse dataset (BP4D-spontaneous) and can be used with Action Units (AUs) too. The first three rows shows the extrapolation along the specified AU. We can also synthesize the expressions by combining different AUs. For example, the last row shows "Happiness/Smile" as a result of combining AU-6 and AU-12.

Table 2. In this table, we show quantitative reconstruction comparisons of our method against SOTA on Facewarehouse dataset.

| Method | $E_{avd}(mm)$ | |
|---|---|---|
| | Mean ↓ | Median ↓ |
| Bilinear [8] | 0.993 | 0.998 |
| FLAME [32] | 0.882 | 0.905 |
| CoMA [41] | 0.825 | 0.811 |
| Jiang et al. [25] | 0.472 | 0.381 |
| Chandran et al. [9] | 0.376 | 0.351 |
| *Ours* | **0.293** | **0.279** |

### 4.4. Shape-texture synthesis of subtle expressions

We examine the potential application in controlling the intensity of expression and action unit for Facescape and BP4D-Spontaneous datasets respectively, with synthetic images using our method. To be specific, given a $z_{id}$ and different intensities of $z_{exp}$ belonging to the same expression or combination of AUs, we can generate the mesh and texture with different intensities corresponding to the $z_{id}$ and $z_{exp}$. This allows us to obtain granular control over the expression intensities of the synthesized faces in Facescape. Fig. 9 shows examples of intensity variation of the texture maps along with the corresponding rendered images. We also extend this framework to incorporate controlling the intensity of AUs using the BP4D-Spontaneous dataset, the results of which are shown in Fig. 10.

To demonstrate the effectiveness of our method, we introduce GANimation [42][2] for comparison, which has shown remarkable achievements in animating 2D images with the facial expression of different intensities. GANimation requires AU intensities corresponding to the rendered images as expression labels. Therefore, a common facial landmark detection tool, Openface[3] was used to detect 17 AUs for each image. However, the facial landmarks detected by Openface are not always accurate, especially for the expression "mouth stretch" in Facescape. Thus, we train our model and GANimation on the rendered FaceScape dataset, excluding the "mouth stretch" expression for a fair comparison. For BP4D-Spontaneous dataset, though 5 AUs are labeled with intensity, it does not meet the requirements for GANimation. Therefore, we still leverage Openface for the rendered images from BP4D-Spontaneous to generate AUs intensities.

For comparison with GANimation, we adopted a pretrained GANimation model and fine-tuned it. Aforemen-tioned rendered images for both datasets and corresponding generated labels of AU intensities are leveraged for fine-tuning. We randomly generated 20 new identities from our method for both datasets and used the rendered neutral images as the source during inference. After generating 17 intensities of AUs for the target images, we used GANimation to transfer the newly generated neutral faces to the target expression with different intensities from 0 to 2 and compared the results with our model.

---
[2]https://github.com/donydchen/ganimation_replicate
[3]https://github.com/TadasBaltrusaitis/OpenFace

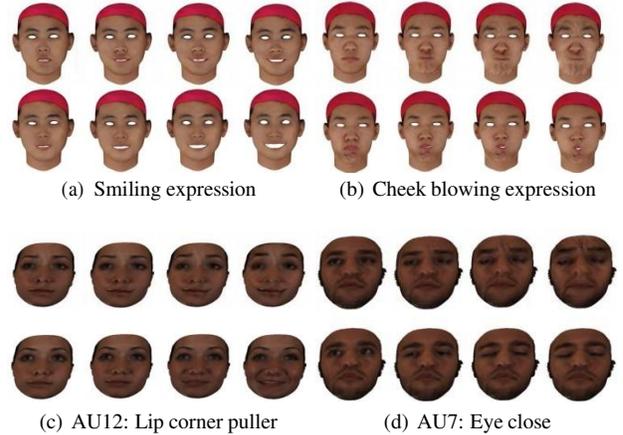

(a) Smiling expression   (b) Cheek blowing expression

(c) AU12: Lip corner puller   (d) AU7: Eye close

Figure 11. (a,b) are results of FaceScape, while (c,d) belongs to BP4D-Spontaneous. In each sub-figure, the first row is the results of GANimation, while the second is ours. From the first to the fourth column, the level of intensities are 0, 1, 1.5, and 2 respectively.

The comparison results are shown in Fig. 11. For the Facescape results, our method can manipulate the expressions better, especially when the level of intensity is beyond 1. For example, in Fig. 11 (a), the mouth opens larger for smiling than GANimation. Another observation is that GANimation can produce noticeable artifacts, as can be seen in Fig. 11 (b). For the BP4D-Spontaneous dataset, our method can also manipulate the level of action unit better. As can be inferred from Fig. 11 (c,d), our method can extrapolate along the AU without artifacts. However, similar to FaceScape, GANimation creates artifacts, and in this case, it sometimes creates faces that are hard to be recognized as human faces. The reason for the difference in performance is that our method can disentangle the identity, and controls the expression geometrically while maintaining 3D consistency.

## 5. Conclusion

We proposed a new framework that uses a pair of supervised Auto-Encoder (SAE) and a cGAN to synthesize high-quality textures with high-frequency details and shapes, delivering granular control over the expressions. The SAE explicitly uses two encoders to non-linearly map the 3D facial meshes into two compact, disentangled identity and expression subspaces in a supervised manner. The two encoders share no parameters, allowing us to decouple the identity and expression factors completely. However, the correlation between identity and expression representations is maintained by sharing the same decoder while reconstructing the original shapes. While the sampling from the disentangled subspaces learned by the SAE space is not trivial, our method uses a cGAN to provide a normalized sampling scheme. Likewise, the framework uses another cGAN trained on high-resolution texture maps as our statistical representation of the facial textures for rendering images.

# References

[1] Victoria Fernández Abrevaya, Adnane Boukhayma, Stefanie Wuhrer, and Edmond Boyer. A decoupled 3d facial shape model by adversarial training. In *Proceedings of the IEEE/CVF International Conference on Computer Vision*, pages 9419–9428, 2019.

[2] Victoria Fernández Abrevaya, Adnane Boukhayma, Stefanie Wuhrer, and Edmond Boyer. A generative 3d facial model by adversarial training. 2019.

[3] Victoria Fernández Abrevaya, Stefanie Wuhrer, and Edmond Boyer. Multilinear autoencoder for 3d face model learning. In *2018 IEEE Winter Conference on Applications of Computer Vision (WACV)*, pages 1–9. IEEE, 2018.

[4] Brian Amberg, Reinhard Knothe, and Thomas Vetter. Expression invariant 3d face recognition with a morphable model. In *2008 8th IEEE International Conference on Automatic Face & Gesture Recognition*, pages 1–6. IEEE, 2008.

[5] Timur Bagautdinov, Chenglei Wu, Jason Saragih, Pascal Fua, and Yaser Sheikh. Modeling facial geometry using compositional vaes. In *Proceedings of the IEEE Conference on Computer Vision and Pattern Recognition*, pages 3877–3886, 2018.

[6] Volker Blanz and Thomas Vetter. A morphable model for the synthesis of 3d faces. In *Proceedings of the 26th annual conference on Computer graphics and interactive techniques*, pages 187–194, 1999.

[7] Timo Bolkart and Stefanie Wuhrer. A robust multilinear model learning framework for 3d faces. In *Proceedings of the IEEE conference on computer vision and pattern recognition*, pages 4911–4919, 2016.

[8] James Booth, Anastasios Roussos, Stefanos Zafeiriou, Allan Ponniah, and David Dunaway. A 3d morphable model learnt from 10,000 faces. In *Proceedings of the IEEE conference on computer vision and pattern recognition*, pages 5543–5552, 2016.

[9] Alan Brunton, Augusto Salazar, Timo Bolkart, and Stefanie Wuhrer. Review of statistical shape spaces for 3d data with comparative analysis for human faces. *Computer Vision and Image Understanding*, 128:1–17, 2014.

[10] Chen Cao, Yanlin Weng, Shun Zhou, Yiying Tong, and Kun Zhou. Facewarehouse: A 3d facial expression database for visual computing. *IEEE Transactions on Visualization and Computer Graphics*, 20(3):413–425, 2013.

[11] Prashanth Chandran, Derek Bradley, Markus Gross, and Thabo Beeler. Semantic deep face models. In *2020 International Conference on 3D Vision (3DV)*, pages 345–354. IEEE, 2020.

[12] Paolo Cignoni, Marco Callieri, Massimiliano Corsini, Matteo Dellepiane, Fabio Ganovelli, Guido Ranzuglia, et al. Meshlab: an open-source mesh processing tool. In *Eurographics Italian chapter conference*, volume 2008, pages 129–136. Salerno, Italy, 2008.

[13] Timothy F. Cootes, Gareth J. Edwards, and Christopher J. Taylor. Active appearance models. *IEEE Transactions on pattern analysis and machine intelligence*, 23(6):681–685, 2001.

[14] Jiankang Deng, Shiyang Cheng, Niannan Xue, Yuxiang Zhou, and Stefanos Zafeiriou. Uv-gan: Adversarial facial uv map completion for pose-invariant face recognition. In *Proceedings of the IEEE conference on computer vision and pattern recognition*, pages 7093–7102, 2018.

[15] Bernhard Egger, William AP Smith, Ayush Tewari, Stefanie Wuhrer, Michael Zollhoefer, Thabo Beeler, Florian Bernard, Timo Bolkart, Adam Kortylewski, Sami Romdhani, et al. 3d morphable face models—past, present, and future. *ACM Transactions on Graphics (TOG)*, 39(5):1–38, 2020.

[16] Yao Feng, Fan Wu, Xiaohu Shao, Yanfeng Wang, and Xi Zhou. Joint 3d face reconstruction and dense alignment with position map regression network. In *Proceedings of the European conference on computer vision (ECCV)*, pages 534–551, 2018.

[17] Xinbo Gao, Ya Su, Xuelong Li, and Dacheng Tao. A review of active appearance models. *IEEE Transactions on Systems, Man, and Cybernetics, Part C (Applications and Reviews)*, 40(2):145–158, 2010.

[18] Baris Gecer, Binod Bhattarai, Josef Kittler, and Tae-Kyun Kim. Semi-supervised adversarial learning to generate photorealistic face images of new identities from 3d morphable model. In *Proceedings of the European conference on computer vision (ECCV)*, pages 217–234, 2018.

[19] Baris Gecer, Alexandros Lattas, Stylianos Ploumpis, Jiankang Deng, Athanasios Papaioannou, Stylianos Moschoglou, and Stefanos Zafeiriou. Synthesizing coupled 3d face modalities by trunk-branch generative adversarial networks. In *European conference on computer vision*, pages 415–433. Springer, 2020.

[20] Baris Gecer, Stylianos Ploumpis, Irene Kotsia, and Stefanos Zafeiriou. Ganfit: Generative adversarial network fitting for high fidelity 3d face reconstruction. In *Proceedings of the IEEE/CVF conference on computer vision and pattern recognition*, pages 1155–1164, 2019.

[21] Jose Gonzalez-Mora, Fernando De la Torre, Rajesh Murthi, Nicolas Guil, and Emilio L Zapata. Bilinear active appearance models. In *2007 IEEE 11th International Conference on Computer Vision*, pages 1–8. IEEE, 2007.

[22] Ian Goodfellow, Jean Pouget-Abadie, Mehdi Mirza, Bing Xu, David Warde-Farley, Sherjil Ozair, Aaron Courville, and Yoshua Bengio. Generative adversarial nets. In *Proc. Neural Information Processing Systems (NIPS)*, pages 2672–2680. 2014.

[23] Ishaan Gulrajani, Faruk Ahmed, Martin Arjovsky, Vincent Dumoulin, and Aaron C Courville. Improved training of wasserstein gans. *Advances in neural information processing systems*, 30, 2017.

[24] Dong Huang and Fernando De la Torre. Bilinear kernel reduced rank regression for facial expression synthesis. In *European conference on computer vision*, pages 364–377. Springer, 2010.

[25] Loc Huynh, Weikai Chen, Shunsuke Saito, Jun Xing, Koki Nagano, Andrew Jones, Paul Debevec, and Hao Li. Mesoscopic facial geometry inference using deep neural networks. In *Proceedings of the IEEE Conference on Computer Vision and Pattern Recognition*, pages 8407–8416, 2018.

[26] Alexandru Eugen Ichim, Sofien Bouaziz, and Mark Pauly. Dynamic 3d avatar creation from hand-held video input. *ACM Transactions on Graphics (ToG)*, 34(4):1–14, 2015.

[27] Zi-Hang Jiang, Qianyi Wu, Keyu Chen, and Juyong Zhang. Disentangled representation learning for 3d face shape. In *Proceedings of the IEEE/CVF Conference on Computer Vision and Pattern Recognition*, pages 11957–11966, 2019.

[28] Tero Karras, Timo Aila, Samuli Laine, and Jaakko Lehtinen. Progressive growing of gans for improved quality, stability, and variation. *arXiv preprint arXiv:1710.10196*, 2017.

[29] Hyeongwoo Kim, Pablo Garrido, Ayush Tewari, Weipeng Xu, Justus Thies, Matthias Niessner, Patrick Pérez, Christian Richardt, Michael Zollhöfer, and Christian Theobalt. Deep video portraits. *ACM Transactions on Graphics (TOG)*, 37(4):1–14, 2018.

[30] Diederik P Kingma and Max Welling. Auto-encoding variational bayes. *arXiv preprint arXiv:1312.6114*, 2013.

[31] Alexandros Lattas, Stylianos Moschoglou, Baris Gecer, Stylianos Ploumpis, Vasileios Triantafyllou, Abhijeet Ghosh, and Stefanos Zafeiriou. Avatarme: Realistically renderable 3d facial reconstruction" in-the-wild". In *Proceedings of the IEEE/CVF conference on computer vision and pattern recognition*, pages 760–769, 2020.

[32] John P Lewis, Ken Anjyo, Taehyun Rhee, Mengjie Zhang, Frederic H Pighin, and Zhigang Deng. Practice and theory of blendshape facial models. *Eurographics (State of the Art Reports)*, 1(8):2, 2014.

[33] Ruilong Li, Karl Bladin, Yajie Zhao, Chinmay Chinara, Owen Ingraham, Pengda Xiang, Xinglei Ren, Pratusha Prasad, Bipin Kishore, Jun Xing, et al. Learning formation of physically-based face attributes. In *Proceedings of the IEEE/CVF conference on computer vision and pattern recognition*, pages 3410–3419, 2020.

[34] Tianye Li, Timo Bolkart, Michael J Black, Hao Li, and Javier Romero. Learning a model of facial shape and expression from 4d scans. *ACM Trans. Graph.*, 36(6):194–1, 2017.

[35] Yaron Lipman, Olga Sorkine, David Levin, and Daniel Cohen-Or. Linear rotation-invariant coordinates for meshes. *ACM Transactions on Graphics (ToG)*, 24(3):479–487, 2005.

[36] Stephen Lombardi, Jason Saragih, Tomas Simon, and Yaser Sheikh. Deep appearance models for face rendering. *ACM Transactions on Graphics (ToG)*, 37(4):1–13, 2018.

[37] Safa Medin, Bernhard Egger, Anoop Cherian, Ye Wang, Joshua Tenanbaum, Xiaoming Liu, and Tim K Marks. Mostgan: 3d morphable stylegan for disentangled face image manipulation. *arXiv*, 2021.

[38] Mehdi Mirza and Simon Osindero. Conditional generative adversarial nets. *arXiv preprint arXiv:1411.1784*, 2014.

[39] Stylianos Moschoglou, Stylianos Ploumpis, Mihalis A Nicolaou, Athanasios Papaioannou, and Stefanos Zafeiriou. 3dfacegan: Adversarial nets for 3d face representation, generation, and translation. *International Journal of Computer Vision*, 128(10):2534–2551, 2020.

[40] Augustus Odena, Christopher Olah, and Jonathon Shlens. Conditional image synthesis with auxiliary classifier gans. In *International conference on machine learning*, pages 2642–2651. PMLR, 2017.

[41] Nicolas Olivier, Kelian Baert, Fabien Danieau, Franck Multon, and Quentin Avril. Facetunegan: Face autoencoder for convolutional expression transfer using neural generative adversarial networks. *arXiv preprint arXiv:2112.00532*, 2021.

[42] A. Pumarola, A. Agudo, A.M. Martinez, A. Sanfeliu, and F. Moreno-Noguer. Ganimation: One-shot anatomically consistent facial animation. 2019.

[43] Anurag Ranjan, Timo Bolkart, Soubhik Sanyal, and Michael J Black. Generating 3d faces using convolutional mesh autoencoders. In *Proceedings of the European Conference on Computer Vision (ECCV)*, pages 704–720, 2018.

[44] Elad Richardson, Matan Sela, and Ron Kimmel. 3d face reconstruction by learning from synthetic data. In *2016 fourth international conference on 3D vision (3DV)*, pages 460–469. IEEE, 2016.

[45] Matan Sela, Elad Richardson, and Ron Kimmel. Unrestricted facial geometry reconstruction using image-to-image translation. In *Proceedings of the IEEE International Conference on Computer Vision*, pages 1576–1585, 2017.

[46] Gil Shamai, Ron Slossberg, and Ron Kimmel. Synthesizing facial photometries and corresponding geometries using generative adversarial networks. *ACM Transactions on Multimedia Computing, Communications, and Applications (TOMM)*, 15(3s):1–24, 2019.

[47] Harry Shum and Sing Bing Kang. Review of image-based rendering techniques. In *Visual Communications and Image Processing 2000*, volume 4067, pages 2–13. SPIE, 2000.

[48] Ron Slossberg, Gil Shamai, and Ron Kimmel. High quality facial surface and texture synthesis via generative adversarial networks. In *Proceedings of the European Conference on Computer Vision (ECCV) Workshops*, pages 0–0, 2018.

[49] Qingyang Tan, Lin Gao, Yu-Kun Lai, and Shihong Xia. Variational autoencoders for deforming 3d mesh models. In *Proceedings of the IEEE conference on computer vision and pattern recognition*, pages 5841–5850, 2018.

[50] Justus Thies, Michael Zollhofer, Marc Stammminger, Christian Theobalt, and Matthias Nießner. Face2face: Real-time face capture and reenactment of rgb videos. In *Proceedings of the IEEE conference on computer vision and pattern recognition*, pages 2387–2395, 2016.

[51] Mukhiddin Toshpulatov, Wookey Lee, and Suan Lee. Generative adversarial networks and their application to 3d face generation: a survey. *Image and Vision Computing*, 108:104119, 2021.

[52] Laurens Van der Maaten and Geoffrey Hinton. Visualizing data using t-sne. *Journal of machine learning research*, 9(11), 2008.

[53] Mengjiao Wang, Yannis Panagakis, Patrick Snape, and Stefanos Zafeiriou. Learning the multilinear structure of visual

data. In *Proceedings of the IEEE conference on computer vision and pattern recognition*, pages 4592–4600, 2017.

[54] Thibaut Weise, Sofien Bouaziz, Hao Li, and Mark Pauly. Realtime performance-based facial animation. *ACM transactions on graphics (TOG)*, 30(4):1–10, 2011.

[55] Shugo Yamaguchi, Shunsuke Saito, Koki Nagano, Yajie Zhao, Weikai Chen, Kyle Olszewski, Shigeo Morishima, and Hao Li. High-fidelity facial reflectance and geometry inference from an unconstrained image. *ACM Transactions on Graphics (TOG)*, 37(4):1–14, 2018.

[56] Haotian Yang, Hao Zhu, Yanru Wang, Mingkai Huang, Qiu Shen, Ruigang Yang, and Xun Cao. Facescape: a large-scale high quality 3d face dataset and detailed riggable 3d face prediction. In *Proceedings of the IEEE/CVF Conference on Computer Vision and Pattern Recognition*, pages 601–610, 2020.

[57] Cha Zhang and Tsuhan Chen. A survey on image-based rendering—representation, sampling and compression. *Signal Processing: Image Communication*, 19(1):1–28, 2004.

[58] Xing Zhang, Lijun Yin, Jeffrey F Cohn, Shaun Canavan, Michael Reale, Andy Horowitz, Peng Liu, and Jeffrey M Girard. Bp4d-spontaneous: a high-resolution spontaneous 3d dynamic facial expression database. *Image and Vision Computing*, 32(10):692–706, 2014.

[59] Michael Zollhöfer, Justus Thies, Pablo Garrido, Derek Bradley, Thabo Beeler, Patrick Pérez, Marc Stamminger, Matthias Nießner, and Christian Theobalt. State of the art on monocular 3d face reconstruction, tracking, and applications. In *Computer Graphics Forum*, volume 37, pages 523–550. Wiley Online Library, 2018.

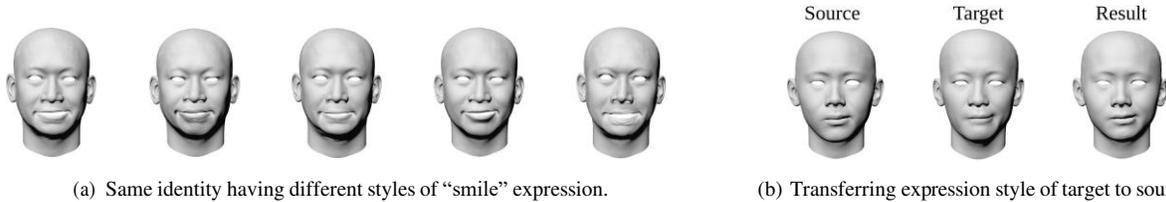

(a) Same identity having different styles of "smile" expression.

(b) Transferring expression style of target to source.

Figure 12. Style editing: changing only the $z_{id}$ code in $G_{exp}$, the style of expression or fine details can be transferred while preserving the identity.

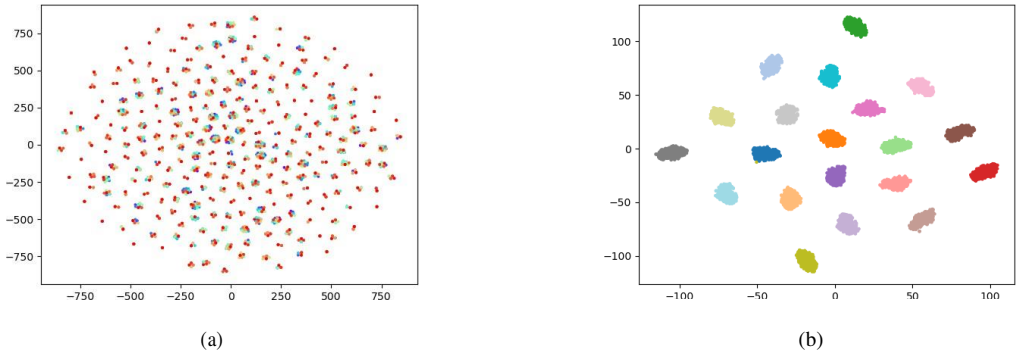

(a)

(b)

Figure 13. (a) and (b) are visualization results of identity and expression embedding respectively of generated samples with same level of expressions using t-SNE. There are 20 colors in each figure indicating the 20 expressions.

## Appendix 1

**Expression generator conditioned on identity:** A commonly acknowledged fact about human faces is each individual's ability to express themselves uniquely with differences in the fine details. The expression generator when trained with inputs as ($z_{noise}$, $z_{exp}$) can generate only a single style of expression for each identity. By using the $G_{exp}$ conditioned on $z_{id}$, our model can modify the style of expression associated with a shape's identity by changing only the $z_{id}$ code in $G_{exp}$, thereby increasing the overall diversity. As illustrated in Fig. 12(a), the same identity has the "smile" expression in various styles. In Fig. 12(b), the source identity has a "mouth right" expression, and the target has the same expression with a slight smile and deeper lip corner, which is transferred onto the source shape.**Visualization of generated samples:** We explore the embedding space of generated samples with the same level of expressions. As shown in Fig. 13, we conduct t-SNE on 5000 samples generated by our model, including 250 different identities with 20 different expressions each. We also explore the embedding space of generated samples with different levels of expressions. Specifically, we generate 3000 samples, including 15 different levels for all 20 expressions, and each level of a specific expression has 10 samples. The intensity of expression varies from 0.0 to 1.5 (higher values of number mean a higher level of intensity of expression). As shown in Fig. 14(a) by t-SNE, all the clusters can be roughly divided into two categories, i.e., ones with pure colors and others with mixed colors. Each color represents a specific expression. We randomly zoom in a cluster with pure color (region A) and the expression levels are primarily high values (above 0.5).

Besides, the samples close to the cluster's center have higher values (the most immediate samples have the highest values of 1.5). This is because samples with higher levels or intensities visually have more prominent expressions, and they are more likely to belong to that corresponding expressions. And so does their embedding space as in Fig. 14(b). The samples with lower levels of expressions, on the other hand, are more like 'neutral' even though they have different expressions. As shown in Fig. 14(c), we randomly zoom in a cluster with mixed color (region B), and the expression levels are mostly low values (lower than 0.5). Besides, the samples close to the cluster's center have lower values (the closest samples have the lowest values of 0.0). This set of experiments shows that our generative model can not only generate different levels of expressions visually but also make sense semantically in the corresponding embedding space. And our model indeed encodes identity and expression information in an implicitly meaningful way.

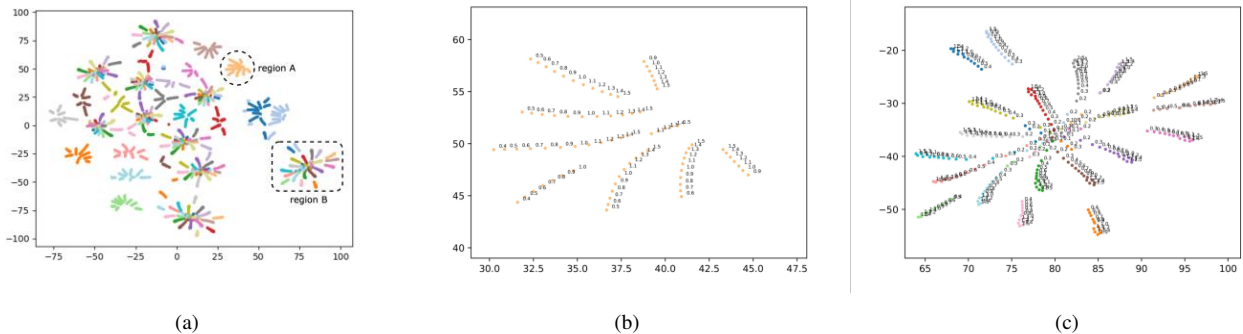

Figure 14. (a) visualization of expression embedding of generated samples with different levels of expressions using t-SNE. (b) zoom of region A of (a) and (c) zoom of region B of (a). The numbers in (b) and (c) denote the different levels or intensities of expressions.

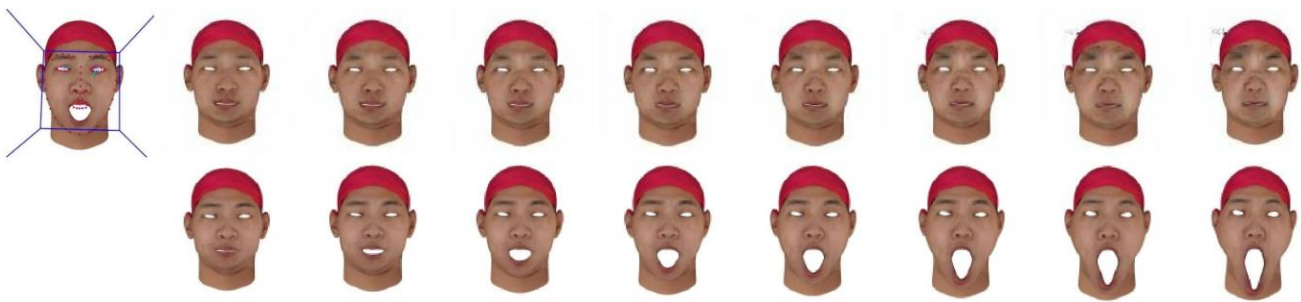

Figure 15. The first row shows the extrapolation results of GANimation when the AUs are incorrect, while the second row shows the extrapolation results by our model.

## Appendix 2

Here we present the results when the AUs are incorrect are shown in Fig. 15. We can observe that GANimation results hardly change since, with the wrong facial landmarks, the model regards the value of action units corresponding to the mouth region as nearly a constant from the source to the target. In this way, the images will not change no matter we perform interpolation or extrapolation.
In conclusion, our method has superior expression intensity control compared with GANimation.
**Architecture details:** In table 3 and table 4 we detail the architecture of our encoder and decoder. Here, *num_verts* denotes the number of a shape's vertices within a dataset and *num_verts* $\times 3$ means the dimension of the flattened vector comprising all vertices' 3D coordinates (*x, y, z*). Besides, $n_{id}$ and $n_{exp}$ are the number of identities and expressions within a dataset, respectively. In particular, *num_verts* = 26317, $n_{id}$ = 847 and $n_{exp}$ = 20 for FaceScape dataset and *num_verts* = 22127, $n_{id}$ = 267 and $n_{exp}$ = 7 for Comb dataset. In table 5 and table 6, we detail the architecture of our generator and discriminator. Identity code $z_{id}$ and noise code $z_{noise}$ are sampled from Gaussian distribution and expression code is a one-hot vector. We use LeakyReLU with a slope of 0.2 as our activation function.

## Appendix 3

Fig. 16 illustrates extrapolation along the textures of randomly generated identities. By fixing the expression code, we can fix the expression and change the identity by varying the identity code and vice versa. Fig. 17 shows the rendered faces with the corresponding meshes for a given ID and different expressions.

We also trained our model on BP4D dataset - a highly diverse dataset. The dataset are labeled by AUs and tasks instead of holistic expression labels.

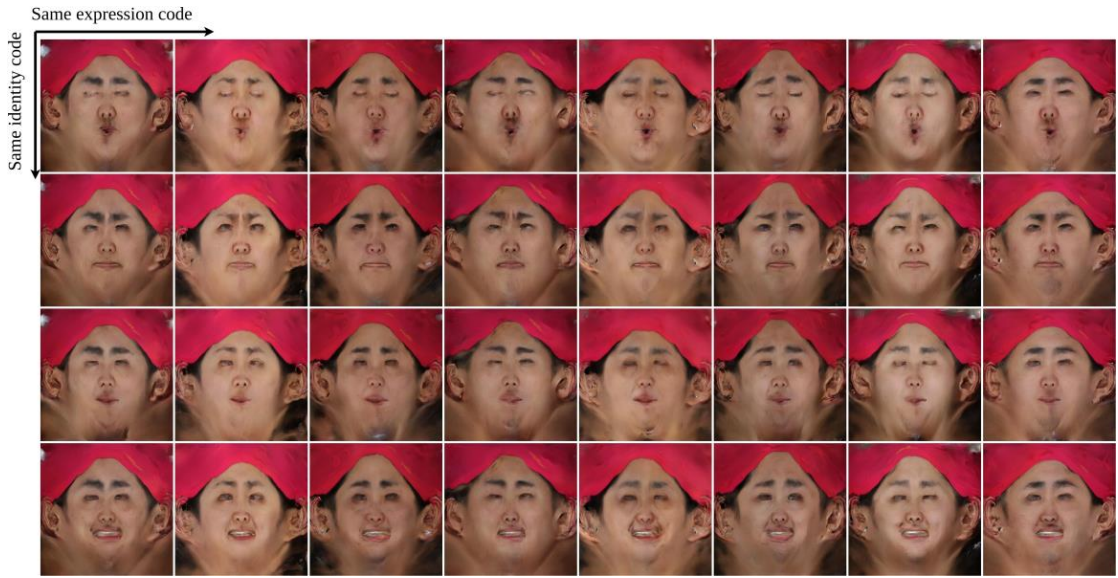

Figure 16. Generated Textures.

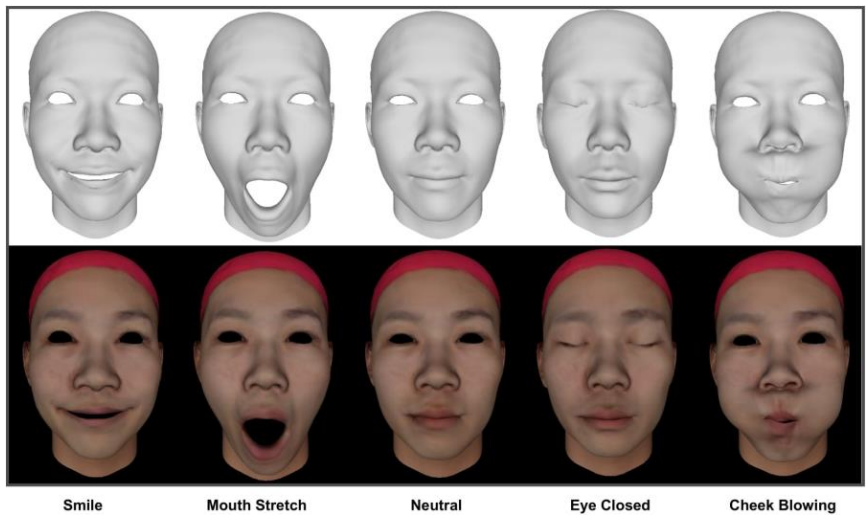

Figure 17. Meshes and rendered images of same ID, different expressions. The first row shows meshes generated using our shape generator and the second row shows the corresponding rendered images using the textures from the texture generator.

Table 3. The architecture of encoder

|  | ID encoder | Exp encoder |
| --- | --- | --- |
| Input | $num\_verts \times 3$ | $num\_verts \times 3$ |
| Linear | $num\_verts \times 3 \rightarrow 1024$ | $num\_verts \times 3 \rightarrow 1024$ |
| Activation | LeakyReLU(0.2) | |
| Linear | 1024→512 | 1024→512 |
| Activation | LeakyReLU(0.2) | |
| Linear | 512→100 | 512→30 |
| Output | ID Embedding | Exp Embedding |
| | **Classification Layer** | **Classification Layer** |
| Linear | $100 \rightarrow n_{id}$ | $30 \rightarrow n_{exp}$ |
| Output | ID class | Exp class |

Table 4. The architecture of decoder

|  | ID decoder | Exp decoder |
|---|---|---|
| Input | ID embedding | Exp embedding |
| Linear | ID embedding → 512 | Exp embedding → 512 |
| Activation | LeakyReLU(0.2) | |
| Linear | 512→1024 | 512→1024 |
| Activation | LeakyReLU(0.2) | |
| Linear | 1024→ *num verts* × 3 | 1024→ *num verts* × 3 |
| Output | *num verts* × 3 | *num verts* × 3 |

Table 5. The architecture of ID generator $G_{id}$, Exp generator $G_{exp}$

|  | $G_{id}$ | $G_{exp}$ |
|---|---|---|
| Input | $z_{id} + z_{noise}$ | $z_{id} + z_{exp} + z_{noise}$ |
| Linear | (20 + 5)→ 512 | (20 + 20 + 5)→512 |
| Activation | LeakyReLU(0.2) | |
| Linear | 512→512 | 512→512 |
| Activation | LeakyReLU(0.2) | |
| Linear | 512→1024 | 512→1024 |
| Activation | LeakyReLU(0.2) | |
| Linear | 1024→100 | 1024→30 |
| Output | ID embedding | Exp embedding |

Table 6. The architecture of discriminator $D$

|  | $D$ |
|---|---|
| Input | ID embedding + Exp embedding |
|  | **Common branch** |
| Linear | (100 + 30)→ 1024 |
| Activation | LeakyReLU(0.2) |
| Linear | 1024→512 |
| Activation | LeakyReLU(0.2) |
| Linear | 512→256 |
| Activation | LeakyReLU(0.2) |
|  | **Discriminator branch** |
| Linear | 256→1 |
| Output | Real or Fake |
|  | **ID branch** |
| Linear | 256→ $n_{id}$ |
| Output | ID class |
|  | **Exp branch** |
| Linear | 256→ $n_{exp}$ |
| Output | Exp class |